%% file: main.tex
\documentclass[conference,a4paper]{IEEEtran}
\IEEEoverridecommandlockouts
\usepackage{cite}
\usepackage{amsmath,amssymb,amsfonts}
\usepackage{algorithmic}
\usepackage{graphicx}
\usepackage{textcomp}
\usepackage{xcolor}
\usepackage{cite}
\usepackage{color}
\usepackage{balance}
\usepackage{booktabs}
\usepackage{subfig}
\usepackage[export]{adjustbox}
\usepackage{mathtools}
\usepackage{mathptmx}
\usepackage{txfonts}

\usepackage{siunitx}

\DeclarePairedDelimiter\floor{\lfloor}{\rfloor}

\usepackage{hyperref}

\def\BibTeX{{\rm B\kern-.05em{\sc i\kern-.025em b}\kern-.08em
    T\kern-.1667em\lower.7ex\hbox{E}\kern-.125emX}}
\begin{document}

\title{Sub-millisecond Video Synchronization \\ of Multiple Android Smartphones
\thanks{The first two authors contributed equally to this work.}
}

\author{\IEEEauthorblockN{
Azat Akhmetyanov\IEEEauthorrefmark{1}\IEEEauthorrefmark{2}, 
Anastasiia Kornilova\IEEEauthorrefmark{1}, 
Marsel Faizullin\IEEEauthorrefmark{1},
David Pozo\IEEEauthorrefmark{1},
and
Gonzalo Ferrer\IEEEauthorrefmark{1}}
\IEEEauthorblockA{\IEEEauthorrefmark{1}Skolkovo Institute of Science and Technology, \IEEEauthorrefmark{2}Saint Petersburg State University \\
Email: {\tt\small \{azat.akhmetyanov,anastasiia.kornilova,marsel.faizullin,d.pozo,g.ferrer\}@skoltech.ru}
}
}

\maketitle

\begin{abstract}
This paper addresses the problem of building an affordable easy-to-setup synchronized multi-view camera system, which is in demand for many Computer Vision and Robotics applications in high-dynamic environments. In our work, we propose a solution for this problem~--- a publicly-available Android application for synchronized video recording on multiple smartphones with sub-millisecond accuracy. We present a generalized mathematical model of timestamping for Android smartphones and prove its applicability on 47 different physical devices. Also, we estimate the time drift parameter for those smartphones, which is less than \SI{1.2}{\milli\second} per minute for most of the considered devices, that makes smartphones' camera system a worthy analog for professional multi-view systems. Finally, we demonstrate Android-app performance on the camera system built from Android smartphones quantitatively on setup with lights and qualitatively~--- on panorama stitching task.
\end{abstract}


\input{src/01_intro}

\input{src/02_related}

\input{src/04_android_phase}
\input{src/05_video_sync}
\input{src/06_conclusion}

\clearpage

\bibliographystyle{bib/IEEEtran}
\balance
\bibliography{bib/IEEEabrv,bib/main}

\end{document}

%% file: src/01_intro.tex
\section{Introduction}

Perception systems that employ synchronized stereo or multi-view camera networks give more opportunities than monocular systems in a wide variety of tasks: depth estimation~\cite{zhang2020du}, 3D reconstruction and rendering~\cite{bortolon2021multi, mildenhall2020nerf}, human pose estimation~\cite{qiu2019cross, dong2019fast, remelli2020lightweight}, especially in high-dynamic scenes. However, to build such a system, additional specialized equipment and expertise are required as well as special tools to provide their synchronization. Those requirements create an abyss between developed algorithms and end-user. 


The active usage of smartphones and the constant growth of their capabilities in terms of sensors quality, camera, Inertial Measurement Unit (IMU), etc. and computational performance make them an affordable tool for usage in such technological pipelines, as it has been incessantly showcased during the last decade in AR/VR applications, robotics~\cite{mueller2021openbot}, and medicine~\cite{kornilova2021deep}. 




Besides time synchronization between devices, moments of camera shots should also be synchronized to achieve better data quality. In professional cameras, it is usually implemented by hardware triggering, unavailable from the smartphone API. To achieve frame-level sync on smartphones, instead of triggering, one can align camera streams from continuous streaming mode. Assuming that stream rate is equal among devices, phase shifts of those periodical signals should be aligned. Existing works propose continuously reloading the camera module until the phase shift will be aligned~\cite{latimer2014socialsync} or requesting frames with different exposure times, which shifts phase to target one in a more controllable and deterministic way~\cite{ansari2019wireless}. Although the purpose of those works is on aligning phase shifts during preview mode and extracting still images from preview, video recording is more advantageous because of hardware codec optimizations (H264, H265) and it provides more frequent and continuous observations.


Our work presents a fully automated Android application for recording synchronized videos and images from multiple Android smartphones with sub-millisecond synchronization accuracy. It is based on the phase shift alignment algorithm from the previous work~\cite{ansari2019wireless} and extends it with analysis of phase shift during switching between preview and video mode. We also formulate a general mathematical model of frame timestamping on Android devices and define an optimization problem for calculating model parameters (period, phase shift). We propose an effective algorithm to estimate those parameters and evaluate the algorithm on 47 physical devices. Synchronization accuracy of implemented Android application is quantitatively evaluated on setup with lights and qualitatively~--- on panorama stitching of a dynamic scene. The app is publicly available\footnote{\url{https://github.com/MobileRoboticsSkoltech/RecSync-android}}.

\begin{figure*}[ht]
    \centering
    \includegraphics[width=0.97\textwidth]{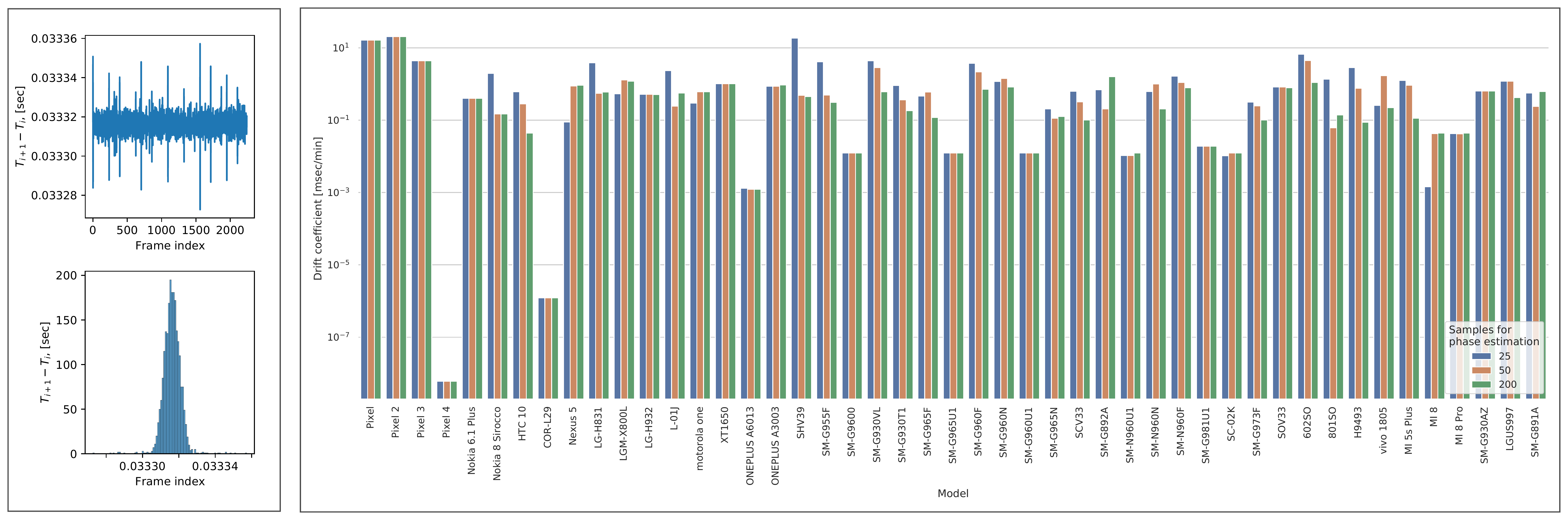}
    \caption{{\em Top-Left:} Difference of consecutive timestamps for Samsung Galaxy S9+. {\em Bottom:Left:} Its histogram resembling a Gaussian. {\em Right:} Drift coefficients (\SI{}{\milli\second} per minute) for 47 different smartphones from Firebase database in logarithmic scale using different amount of samples for phase estimation.}
    \label{fig:drift}
    \vspace{-4mm}
\end{figure*}

%% file: src/02_related.tex
\section{Related work}

Camera synchronization of multiple smartphones deals with two main issues: clock synchronization in sensor networks and camera phase alignment. Approaches for clock synchronization could be divided into the following groups: (1) hardware synchronization using wires and special boards~\cite{tschopp2020versavis, liu2021matter}, usually not applicable to smartphones, (2) network-based approaches in which sensor nodes exchange timing messages via the network and aggregate statistics (NTP~\cite{rfc4330}, PTP~\cite{ptp2008}, RBS~\cite{rbs2006}), (3) data-based approaches based on aligning time of some event registered by data from all sensors~\cite{sandha2019exploiting}, for example, flashlights~\cite{vsmid2017rolling}, sound~\cite{ono2016self, sandha2019exploiting}, shake~\cite{faizullin2021twist}. Besides the topic of camera phase synchronization is well investigated in the professional equipment domain, the smartphone case is complicated by a variety of models and high-level API and was studied only in a couple of works. Among camera phase alignment approaches, SocialSync~\cite{latimer2014socialsync} and libsoftwaresync~\cite{ansari2019wireless} could be highlighted that employ different techniques for shifting camera phase.


As the basis of our work, we use the libsoftwaresync approach of synchronized photo capturing with sub-milliseconds accuracy, which consists of two steps: (1) perform clock synchronization between smartphones using min-filtering NTP modification, (2) align camera phases of all devices to a common one using additional requests of frames with different exposure to shift the phase during preview mode.

There are \textbf{two open questions} to which our research tries to give answers: (1) could this approach be extended to video recording (which is equivalent to phase stability after switch from preview to video mode) and (2) could this approach be scaled to the whole Android smartphone family and what is the amount of drift for common devices.

%% file: src/04_android_phase.tex
\section{Camera phase analysis}
\label{sec:requirements}
In our work, we consider a system that contains multiple Android smartphones of the same model with Camera2API support and precise timestamping of frames that is guaranteed by properties of the camera metadata~\cite{cameradoc}. We assume the task of clock synchronization between smartphones to be solved as it was demonstrated in \cite{ansari2019wireless, faizullin2021twist} and focus on phase shift stability during switch to video recording.


\subsection{Problem formulation}

For each smartphone in multi-view camera setup, we solve the following problem independently. The instant of time $t_i$ is the {\em observed} moment when an image is captured, according to the camera timestamp. One can define a model of periodical timestamping, similarly as proposed in \cite{ansari2019wireless}:
\begin{equation}
    t_i = \tau_0 + N_i \cdot \tau + \eta_i, \label{eq_timestamping}
\end{equation}
where $\tau_0\in \mathbb{R}$ is the phase shift of the $i$th captured frame $i=1,\ldots,s$ and $\tau \in \mathbb{R}$ is the constant period between consecutive frames captured by the camera module. The integer $N_i \in \mathbb{N}$ is the frame index, which considers potential frame drop ($i\neq N_i$) and $\eta_i \sim p(\tau_0,\tau, N_i)$ is a random variable distributed according to some Probability Density Function (PDF). 


Consequently, the joint estimation problem for the phase shift $\tau_0$, the period $\tau$, and each of the integers in $\{N_i\}^{i=1, ...s}$ can be described as a constrained least-squared optimization model:
\begin{align}
    \min_{\tau_0, \tau, N_1,.., N_s} \sum_{i=1}^{s} (\tau_0 + N_i \cdot \tau - t_i)^2 \label{eq_problem_raw}\\
    \text{s.t.} \quad \tau\left(N_i - 0.5\right) \leq t_i \leq \tau\left(N_i + 0.5\right). \label{eq_constr}
\end{align}

The objective function represents the minimum square error between timestamp $t_i$ and timestamping model \eqref{eq_timestamping}. The constraint \eqref{eq_constr} depicts the timestamping bounds.
Due to the integer nature of $N_i$, this problem is classified as Mixed-Integer Problem (MIP) model. Off-the-shelf solvers, such as Gurobi or CPLEX, are the state-of-the-art tools for providing solutions to MIP problems. However, in practice, such solvers are not adequate for online decision-making \cite{bertsimas2019online} and not suited to be embedded into the mobile phone. 
The interest in this type of problem has been materialized in developing initiatives for creating a new family of solvers for MIP models that require solving in short periods \cite{miosqp}.


\begin{figure}[h]
    \centering
    \includegraphics[width=0.8\columnwidth]{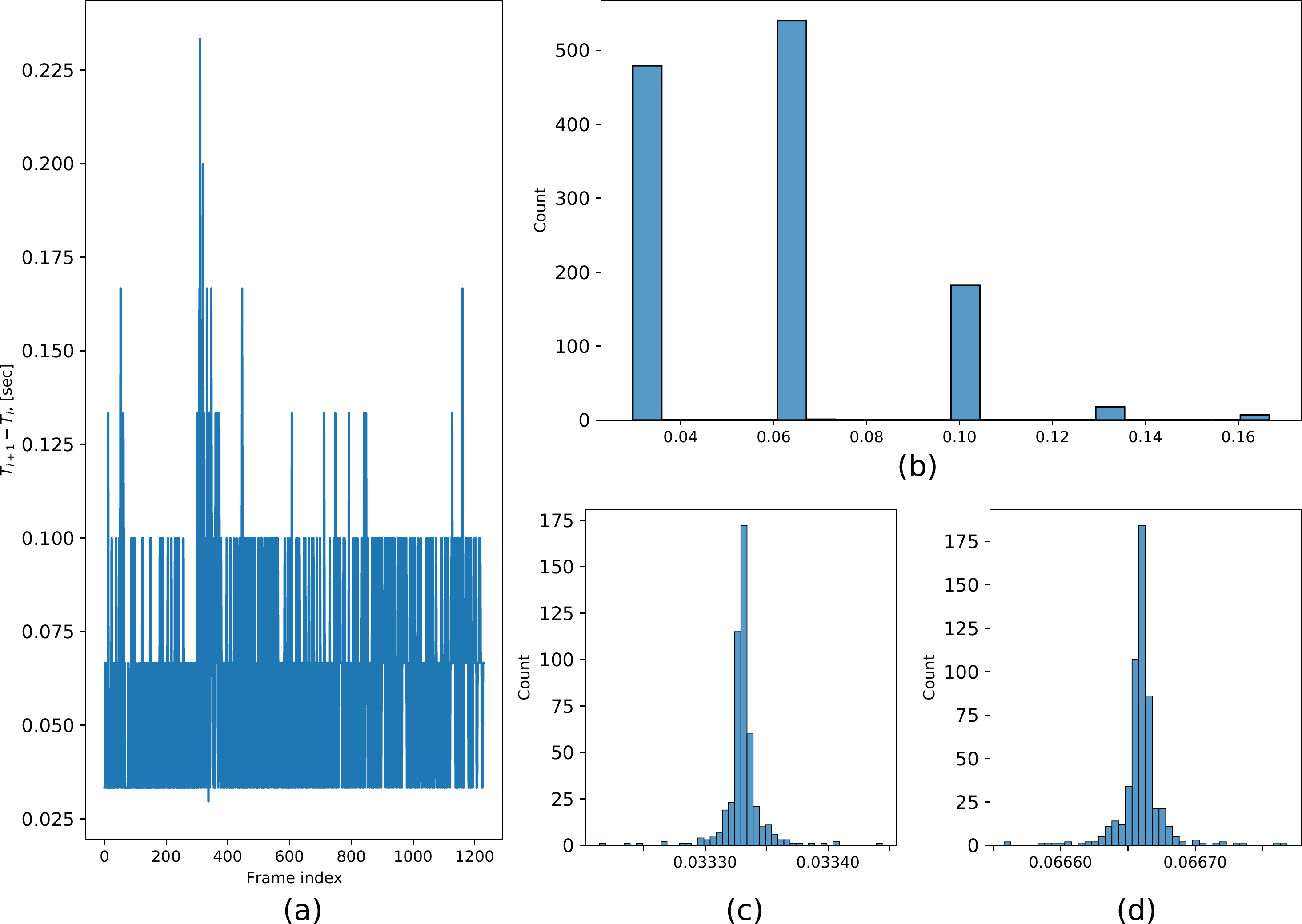}
    \caption{{\em Left:} Difference of consecutive timestamps for HTC 10. {\em Top-Right:} Its histogram showing different modes. {\em Bottom-Right:} Histogram of clustered sequences, now Gaussian.}
    \label{fig:htc}
\end{figure}


The optimization model \eqref{eq_problem_raw}--\eqref{eq_constr}  can be {\em exactly} reformulated into an unconstrained optimization problem. By manipulating the constraint inequalities in (\ref{eq_constr}), it is possible to obtain the exact equality for $N_i$,
\begin{equation}
    N_i(\tau) =  \floor*{\frac{t_i}{\tau} + 0.5},
    \label{eq_round}
\end{equation}
where the operator $\floor*{.}$ indicates the floor operator.









\subsection{Approximated problem reformulation}\label{sec_phase_alg}


The problem statement in \eqref{eq_problem_raw}--\eqref{eq_constr} could be approximated by estimating the $N_i$ and removing its dependency on $\tau$ by: 
\begin{align}
    N_i \approx \floor*{\frac{t_i}{\tau_{init}} + 0.5}, \quad \text{s.t.} \quad
    \tau_{init} = \min_{i} \Delta t_i.
\end{align}

One can further simplify the problem \eqref{eq_problem_raw} by considering finite differences between two consecutive image captures \eqref{eq_finit}:
\begin{equation}
    \Delta t_i = t_i - t_{i - 1} = (N_i - N_{i - 1}) \tau + \eta_i - \eta_{i-1} = \Delta N_i \tau + \xi_{i},\label{eq_finit}
\end{equation}
where we {\em assume} that the new random variable $\xi_i$ is distributed according to normal distribution $\xi_i \sim \mathcal{N}(0, \sigma(\Delta N_i, \tau))$, as discussed in Sec.~\ref{sec_noise}. In that case, the minimization problem could be formulated as in \eqref{eq:delta_min_problem} and could be decomposed into the sum of Least Square (LSQ) problems over the clusters associated with values of $\Delta N_i$. In that case, for every subproblem, the Gaussian assumption over the noise is satisfied therefore every subproblem could be solved using standard methods using the Gauss-Markov theorem. Finally, $\tau$ is calculated as a weighted sum of the subproblem's solutions. 

\begin{equation}
    \label{eq:delta_min_problem}
    \min_{\tau} \sum_{i=1}^{s} (\Delta N_i \tau - \Delta t_i)^2 = \min_{\tau} \sum_{k=1}^{\max N_i} \sum_{\Delta t_i \,\, \text{s.t.} \,\, k = N_i} (k \tau - \Delta t_i)^2
\end{equation}


\subsection{Noise Model}\label{sec_noise}

In this section, we justify the choice for a Gaussian distribution which resulted in the estimation problem in \eqref{eq:delta_min_problem}. Among timestamping data from 47 smartphone models, we observed two classes of timestamping model. On the first kind, finite difference of timestamps follows a Gaussian distribution, as seen in Fig.~\ref{fig:drift}. On the second kind, the smartphone occasionally drops some frames, not shifting the phase but breaking the Gaussian assumption presenting a set of clusters (Fig.~\ref{fig:htc} top-right). Analyzing distribution of every cluster separately, as demonstrated in Fig.~\ref{fig:htc} bottom-right, it could be stated that clusters follow Gaussian distribution. Such decomposition into clusters is done as described in \eqref{eq:delta_min_problem}.

\subsection{Evaluation of timestamping model} 
For analysis of timestamping model on different Android smartphones, we used Firebase~\cite{firebase} platform that provides application's run and test on physical devices. 47 smartphone models of available smartphones were chosen which cameras satisfy the criteria mentioned in Sec.~\ref{sec:requirements}. For every smartphone, camera timestamps were collected during \SI{1}{\minute} of continuous camera operation~--- \SI{15}{\second} of preview mode, \SI{45}{\second} of video mode ($\approx$ 1800 timestamps per device in total). 


To evaluate both phase shift stability after switching from preview mode to video mode and drift of the phase shift over time, we use the drift coefficient metric, which measures drift over time. Phase parameters were estimated using the first 25, 50, 200 observations (``train'' set) from the preview mode. Drift coefficient was estimated on the last 1000 (``test'' set) estimations that contain preview mode, switch to video mode, and video mode, their values are depicted in Fig.~\ref{fig:drift} (right). The majority of models, except early Google Pixel models, have drift coefficient less than \SI{1.2}{\milli\second} per minute on the ``test'' set of timestamps, that allows to make the next conclusions: (1) phase is not shifted during switch from preview to video mode, allowing to use phase alignment for video recording on wide range of smartphones, (2) drift coefficient on majority of smartphones is low enough and could be a worthy analogue to professional systems. Analysis of different ``train'' set size [25; 50; 200] demonstrates that 50 timestamp observations ($\approx$ \SI{2}{\second}) are the optimal condition for phase parameters estimation, lower values lead to the increase of drift coefficient. 

%% file: src/05_video_sync.tex
\section{Experiments}

Evaluation on drift and phase shift stability in the previous section demonstrated that switching camera operating mode from preview to video doesn't shift aligned phases. That means that the libsoftwaresync~\cite{ansari2019wireless} approach on phase shift alignment during preview and then capturing standalone preview photos could be scaled to video recording. The original app was extended with a video recording feature, the automatic algorithm for phase calculation.

\textbf{Light scenes} To evaluate the Android application for synchronized video recording, we used a rig with two Samsung Galaxy S20 smartphones and a light setup described in~\cite{faizullin2021twist}, based on periodical flash blinking and rolling shutter camera property. Five pairs of synchronized videos with a duration of \SI{1}{\minute} were recorded. Then, for every pair of videos, the time difference of capturing light between corresponding synchronized frames was calculated, time difference between corresponding frames among all videos does not exceed \SI{250}{\micro\second}.

\textbf{Panorama stitching} To qualitatively demonstrate synchronization quality, we applied panorama stitching for synchronized stereo video in smartphones on the dynamic scene with a jumping person. CLI interface of Hugin stitching software was used to stitch frame pairs. Stitched video is available via the link \url{https://youtu.be/2AEmuIt5TZc}.

%% file: src/06_conclusion.tex
\section{Conclusion}
This paper answers two tackled research questions. Firstly, phase synchronization could be extended from preview mode to video recording mode without losing synchronization. Secondly, this approach could be scaled to a wide family of Android devices with synchronization drift less than \SI{1.2}{\milli\second} per minute, as we demonstrated on 47 smartphone models. Proposed Android-app was quantitatively and qualitatively verified on the multi-view camera system with Samsung Galaxy S20 smartphones. As a result, we obtained a fully automated publicly-available tool for recording synchronized videos on multiple Android smartphones.

